\documentclass{article} 
\usepackage{iclr2022_conference,times}

\usepackage{amsmath,amsfonts,bm}

\def\eqref#1{equation~\ref{#1}}

\def\1{\bm{1}}

\DeclareMathAlphabet{\mathsfit}{\encodingdefault}{\sfdefault}{m}{sl}
\SetMathAlphabet{\mathsfit}{bold}{\encodingdefault}{\sfdefault}{bx}{n}

\newcommand{\R}{\mathbb{R}}

\newcommand{\normltwo}{L^2}

\usepackage{hyperref}
\usepackage{url}
\usepackage{graphicx}
\usepackage{subfigure}
\usepackage{siunitx}
\usepackage{booktabs}

\newcommand{\Zt}{\mathbf{Z}}
\newcommand{\Ht}{\mathbf{H}^{(t)}}
\newcommand{\Ot}{\mathbf{O}^{(t)}}

\newcommand{\mathsc}[1]{{\normalfont\textsc{#1}}}

\title{Learning heuristics for A*}

\author{Danilo Numeroso \\
University of Pisa \\ 
\texttt{danilo.numeroso@phd.unipi.it} \\
\And
Davide Bacciu \\
University of Pisa\\
\texttt{davide.bacciu@unipi.it} \\
\AND
Petar Veli\v{c}kovi\'{c} \\
DeepMind \\
\texttt{petarv@deepmind.com}
}

\iclrfinalcopy 
\begin{document}

\maketitle

\begin{abstract}
Path finding in graphs is one of the most studied classes of problems in computer science. In this context, search algorithms are often extended with heuristics for a more efficient search of target nodes. In this work we combine recent advancements in Neural Algorithmic Reasoning to learn efficient heuristic functions for path finding problems on graphs. At training time, we exploit multi-task learning to learn jointly the Dijkstra's algorithm and a {\it consistent} heuristic function for the A* search algorithm. At inference time, we plug our learnt heuristics into the A* algorithm. Results show that running A* over the learnt heuristics value can greatly speed up target node searching compared to Dijkstra, while still finding minimal-cost paths.
\end{abstract}

\section{Introduction}
Search algorithms are central to a variety of domains, such as Robotics \citep{mac2016}, Computer Science \citep{gebser2013domain, chen2020retro} and Chemistry \citep{yeh2012}. Efficient search algorithms are usually equipped with heuristic functions to steer the exploration towards the solution. A widely popular algorithm in path finding problems is A* search \citep{hart1968}, whose effectiveness often relies on a careful design of heuristics specialised for the problem at hand \citep{khalidi2020}. 
Such hand-crafted heuristics require domain knowledge, and their design process is often prone to heavy trial-and-error phases. This motivates the recent exploration of learning-based solutions targeting learning of heuristic functions from data. Prior attempts have explored supervised learning \citep{us2013learning, wilt2015building, kim2020learning} as well as imitation learning \citep{pandylearning} to learn heuristics for search algorithms.
In this work, we build on recent advancements in Neural Algorithmic Reasoning \citep{DBLP:journals/patterns/VelickovicB21} to learn heuristics for the A* search algorithm. Algorithmic reasoning in neural networks is an emerging research topic targeting the combination of the adaptivity of learnt models with the robustness and generality of classical algorithmic solutions. Here, we set up an unsupervised learning problem to learn a performant heuristic function for path finding problems on graphs, seeking to combine the algorithmic guarantees of A* with a neural-based approach to automatically extract heuristic values for graph nodes. We specifically target learning of a heuristic that can generalise well in out-of-distribution scenarios. To this end, we seek to regularise our model towards learning a heuristic function which is computed in an "algorithmic" fashion. To do that, we leverage an approach that has previously been found useful to increase generalisation \citep{xhonneux2021transfer}, and that entails joint learning of: (i) a known algorithm for shortest path, i.e. Dijsktra's algorithm \citep{dijkstra1959note}, and (ii) a heuristic function for A*. 

\section{Problem statement}
We study the problem of learning an {\it admissible} and {\it consistent} heuristic function on graphs for A*. In particular, A* solves the shortest path problem between a source node $s$ and target $t$, improving directly on Dijkstra's algorithm by using a heuristic function $h$ to guide its search. Such heuristics may have different properties: (i) {\it admissibility}, if for any node $v$, $h(v)$ never overestimates the real cost of reaching the target node $t$ from $v$; (ii) {\it consistency}, if $\forall (u, v) \in E : h(u) \leq w_{uv} + h(v)$, where $w_{uv}$ is the cost of the edge $(u, v)$. It follows that a consistent heuristic is always admissible too.
An admissible heuristic is a sufficient condition for the optimality of the A* algorithm, i.e. A* is guaranteed to return the optimal path from $s$ to $v$. Additionally, a consistent heuristic ensures that every node in the graph is evaluated at most once while computing the optimal path.

We consider graphs $G = (V, E, \mathbf{X}, \mathbf{A})$, with $\mathbf{X} = \{\mathbf{x}_v \mid \forall v \in V\}$ and $\mathbf{A} = \{\mathbf{a}_{uv} \mid \forall (u,v) \in E\}$ being the set of node and edge features, respectively. To effectively learn an efficient heuristic function on these graphs, we rephrase the learning problem as a constrained maximisation problem w.l.o.g:
\begin{align} \label{eq:objective}
    \max \, & h(t) - h(s) \\
    \text{subject to} \,\, & h(v) - h(u) \leq w_{uv} \,\, \forall (u, v) \in E.
\end{align}
Such formulation leads the learnt heuristic function towards assigning high values for the target node $t$ and nodes close to it, while assigning low scores to nodes distant to $t$ and the lowest value to $s$.
Therefore, in this setting the heuristic function should not be interpreted as an estimated distance from a given node to the target, but rather as a measure of how good it is to include that node in the path. Furthermore, constraint satisfaction serves to the purpose of consistent heuristics.

A key challenge is that the learnt heuristic function needs to generalise on previously unseen, and possibly differently sized, graphs. To enable such generalisation, we exploit the neural algorithmic reasoning blueprint presented in \citep{DBLP:journals/patterns/VelickovicB21}. In doing that, we attempt to learn a heuristic function that is computed in an algorithmic fashion, i.e. making discrete decisions over neighbourhoods. Furthermore, as shown in prior work \citep{xhonneux2021transfer}, multi-task learning over algorithms has beneficial effects on generalisation. Motivated by this, we choose to learn jointly a known algorithm for shortest path on graphs, e.g. Dijkstra's algorithm, and an optimal assignment to \eqref{eq:objective}.

\section{Methodology} \label{sec:methodology}
We follow the encode-process-decode framework presented in \citep{DBLP:conf/cogsci/HamrickABZMTB18} for neural execution of algorithms. Specifically, for each step $t$ of a generic algorithm $A$, the neural computation is decomposed into three main components. 
First, an encoder layer $f_A$ is applied to project the set of algorithm-specific node features $\mathbf{X}$ and edge features $\mathbf{A}$ to a latent space, producing a set of encoded node features $\Zt_V = \{\mathbf{z}_v \mid \forall v \in V \}$ and encoded edge features $\Zt_E = \{\mathbf{z}_e \mid \forall e \in E \}$. These latent representations are then passed into a processor network $P$, which computes step-wise latent node representations $\Ht = P(\Zt_V, \Zt_E, \mathbf{H}^{(t-1)})$,
where $\mathbf{H}^0$ is initialised as the set of null-vectors, i.e. $\mathbf{H}^0 = \{ \mathbf{0} \mid \forall v \in V\}$. 
Finally, a decoder network $g_A$ implements a mapping from the processor latent space to the actual output space of the algorithm, obtaining node-level outputs $\Ot = g_A(\Zt_V, \mathbf{H}^{(t-1)}, \Ht)$.
In order to learn the heuristic function, we include an additional decoder $g_h$ which learns to predict the heuristic values $\mathbf{y}^{(t)} = \{y^{(t)}_v \mid \forall v \in V\}$ from the latent configurations, with $y^{(t)}_v \in \R$.

In our experiments, algorithm $A$ is the Dijkstra's algorithm. Therefore, the node-level outputs $\Ot$ correspond, at each step $t$, to the predicted {\it predecessor} nodes $p^{(t)}_v$ in the shortest path. In other words, $p^{(t)}_v$ indicates which edge to include in the shortest path to reach the node $v$. The heuristic values $\mathbf{y}^{(t)}$, instead are trained to optimise \eqref{eq:objective}. Similarly to prior work \citep{DBLP:conf/iclr/VelickovicYPHB20}, we choose to implement the encoding and decoding functions as linear transformations. Finally, we use a deep graph network \citep{DBLP:journals/nn/BacciuEMP20} as the processor $P$ with Message-Passing Neural Network (MPNN) \citep{DBLP:conf/icml/GilmerSRVD17} convolution and {\sc max} aggregator, in order to take the graph structure and edge information into account.

To learn jointly the predecessor nodes for Dijkstra and the heuristic function, we train $\Ot$ and $\mathbf{y}^{(t)}$ with two different loss functions. The error signal for the former is the cross-entropy loss, supervised on the ground truth Dijkstra's outputs at each step $t$. This means that the model is supervised not only on the final output, but also on all the intermediate steps made by the algorithm. The latter, instead, is trained in an unsupervised fashion, optimising the objective described in \eqref{eq:objective}. For learning purposes, we optimise a relaxation of the constrained problem, making the loss function for $\mathbf{y}^{(t)}$ unconstrained by replacing the constraints with penalties corresponding to the magnitude of each constraint violation. As the main objective $\{h(t) - h(s)\}$ in \eqref{eq:objective} is unbounded, we add an additional penalty term proportional to the squared $\normltwo$-norm of the predicted heuristics, i.e. $||\mathbf{y}^{(t)}||^2$, and a weighing hyperparameter $\lambda$. Similarly to $\Ot$, we also train heuristic values for each step $t$, although we always rely on the heuristics obtained by a 1-step inference at test time, i.e. $\mathbf{y}^{(1)}$. This means that we add an additional overhead to shortest path finding compared to vanilla Dijkstra, due to the cost of computing heuristic values for all nodes. As each node considers its neighbourhood when computing its value, the additional overhead is quantified as $O(|V| \cdot \bar{n})$, where $\bar{n}$ is the average node degree. We refer to the appendix for detailed definition of the model equations and loss function.

\section{Experimental Evaluation}
\begin{table} 
    \small
    \centering
    \begin{tabular}{lllll}
     \toprule
     \bf{Metrics} & \emph{32 nodes} & \emph{96 nodes} & \emph{192 nodes} & \emph{256 nodes}\\
     \midrule
     & \multicolumn{3}{c}{\sc sparse $(p=\frac{log |V|}{|V|})$} \\ 
     \multicolumn{4}{l}{\textit{A*}} \\ 
     Constraints & $99.4\% \pm 0.06\%$ & $99.7\% \pm 0.04\%$ & $99.8\% \pm 0.03\%$ & $99.8\% \pm 0.03\%$ \\
     Relative Distance & $0.03\% \pm 0.00\%$ & $0.18\% \pm 0.05 \%$ & $0.19\% \pm 0.09\%$ & $0.09\% \pm 0.01\%$ \\
     \# iterations & $16.76 \pm 0.22$ & $42.97\pm 0.93 \%$ & $74.00 \pm 2.02$ & $99.4 \pm 2.71$ \\
     Speedup & 0.94 & 1.08 & 1.19 & 1.23\\
     \midrule
     \multicolumn{4}{l}{\textit{Dijkstra}} \\ 
     \# iterations & 19.23 & 53.68 & 97.18 & 132.07 \\
     \midrule
     \midrule
     & \multicolumn{3}{c}{\sc dense $(p=0.35)$} \\
     \multicolumn{4}{l}{\textit{A*}} \\ 
     Constraints & $99.2\% \pm 0.08\%$ & $99.4\% \pm 0.09\%$ & $99.5\% \pm 0.09\%$ & $99.5\% \pm 0.09\%$ \\
     Relative Distance & $0.98\% \pm 0.16\%$ & $10.75\% \pm 0.77 \%$ & $33.5\% \pm 1.79\%$ & $52.8\% \pm 4.55\%$ \\
     \# iterations & $11.73 \pm 0.31$ & $21.92 \pm 1.95 \%$ & $21.85 \pm 3.41$ & $21.44 \pm 4.57$ \\
     Speedup & 0.93 & 2.29 & 4.30 & 4.83\\
     \midrule
     \multicolumn{4}{l}{\textit{Dijkstra}} \\ 
     \# iterations & 17.35 & 50.82 & 96.55 & 132.50 \\
     \midrule
     \midrule
     & \multicolumn{3}{c}{\sc very dense $(p=0.5)$} \\
     \multicolumn{4}{l}{\textit{A*}} \\ 
     Constraints & $99.1\% \pm 0.08\%$ & $99.4\% \pm 0.11\%$ & $99.5\% \pm 0.11\%$ & $99.5\% \pm 0.11\%$ \\
     Relative Distance & $0.7\% \pm 0.15\%$ & $19.8\% \pm 1.53 \%$ & $59.4\% \pm 5.00\%$ & $73.5\% \pm 6.75\%$ \\
     \# iterations & $11.01 \pm 0.34$ & $15.64 \pm 1.60 \%$ & $16.02 \pm 3.37$ & $15.63 \pm 3.85$ \\
     Speedup & 1.22 & 2.60 & 5.06 & 5.24\\
     \midrule
     \multicolumn{4}{l}{\textit{Dijkstra}} \\ 
     \# iterations & 17.17 & 49.85 & 90.42 & 130.40 \\
     
    \bottomrule
    \end{tabular}
    \caption{Performance of A* with learnt heuristic function on 4 out of the 9 test sets, i.e. 32, 96, 192, 256 nodes. We present performance on constraints satisfaction, relative distance and number of iterations over all three datasets (sparse, dense and very dense) averaged across 5 trials. We additionally report the speedup with respect to the execution times of A* and Dijkstra and the average number of iterations needed by Dijkstra to retrieve the shortest path to compare the efficiency of our learner. Note that higher values in Constraints and Speedup corresponds to better performance. Conversely, better performance for Relative Distance and \# iterations corresponds to lower values.}
    \label{tab:results-t1}
\end{table}
\begin{figure}
    \centering 
    \subfigure[A* shortest path accuracy]{
        \centering
        \includegraphics[width=.45\linewidth]{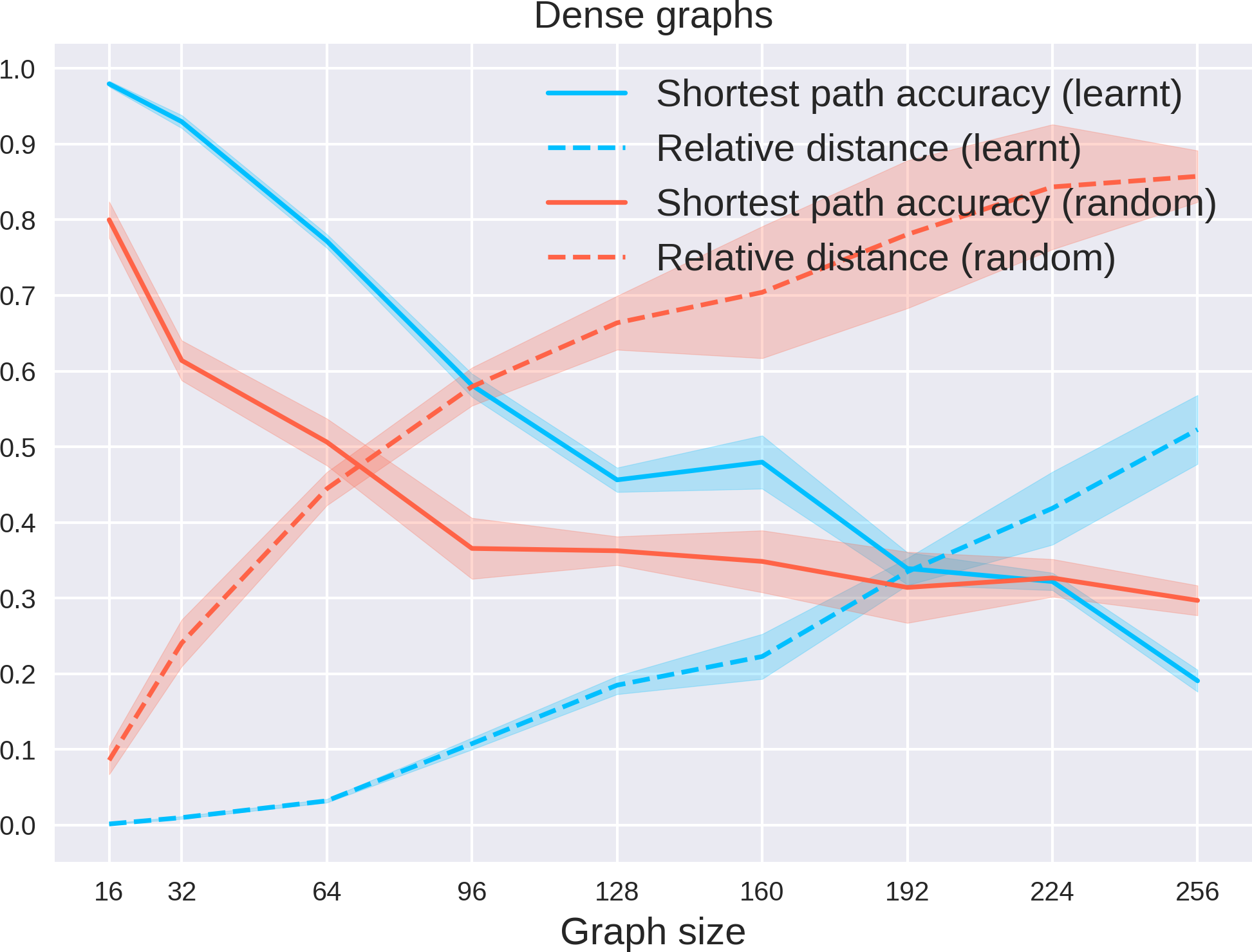}
    }
    \subfigure[A* number of iterations]{
        \centering
        \includegraphics[width=.45\linewidth]{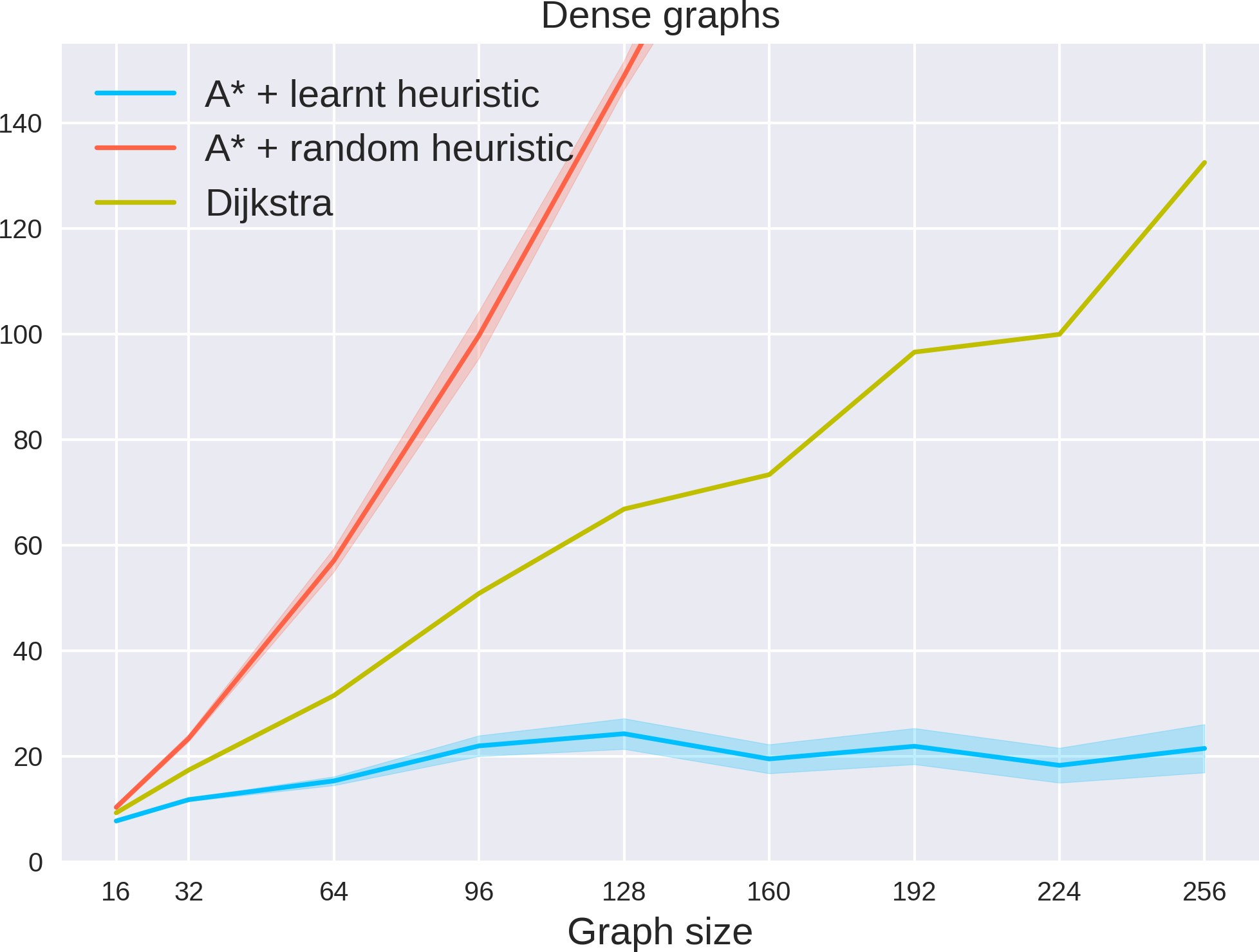}
    }
    \caption{{\bf (a).} Accuracy of A* retrieving the shortest path in the graph solely based on the learnt heuristic function, with respect to different graph sizes. Additionally, we measure the relative distance in terms of cost between the path found by A* and the optimal one, found by executing Dijkstra; {\bf(b).} Number of iterations needed (on average) for A* to reach the target node. We compare the efficiency against a random baseline and Dijkstra; Shaded area represents standard error.}
    \label{fig:results-dense}
\end{figure}

We generate synthetic graphs from the {\it Erdős–Rényi} distribution \citep{erdHos1960evolution} with varying density, to allow training and testing our model on graphs of varying degree of sparsity and size. In particular, we generate three different set of graphs: {\it very dense}, {\it dense} and {\it sparse} graphs,
by setting the probability of sampling an edge between two nodes to $p=0.5$, $p=0.35$ and $p=\frac{\log |V|}{|V|}$, respectively. For each set, we generate small training and validation graphs comprising 16 nodes, i.e. 1000 instances for training purposes and 128 structures for validation. To thoroughly assess the generalisation capability with respect to graph size, we generated 9 different test sets -- 128 graphs each -- with increasing number of nodes, namely 16, 32, 64, 96, 128, 160, 192, 224, 256 nodes. 
In order to train the network on intermediate steps, we utilise the CLRS benchmark \citep{deepmind2021clrs} to generate training data.

We train our model over the set of {\it dense} graphs only, and select the best hyperparameters by a two-level random search with $n=200$ samples. We refer the reader to the appendix for detailed information on the model selection.
Finally, we test our model by running A* with learnt heuristics on all 9 test sets of all 3 datasets, and average the results across 5 trials. 
Results of our experimental analysis are summarised both graphically (for {\it dense} graphs) in Fig. \ref{fig:results-dense} and in tabular form, Table \ref{tab:results-t1}. We report four different metrics: (i) constraints satisfaction accuracy, i.e. how many constraints in \eqref{eq:objective} are satisfied by the learnt assignment; (ii) accuracy with respect to the optimal path $\pi_*$ found by Dijkstra; (iii) average relative distance between the actual retrieved path $\pi$ and $\pi_*$, computed as $\frac{c(\pi) - c(\pi_*)}{c(\pi_*)}$. Here, $c(\cdot)$ represents the cost of the path (summation over edge weights). Specifically, this metric is needed as whenever our learning model leads A* towards finding a sub-optimal path $\pi$, the cost of $\pi$ can either be much higher of $\pi_*$ or can approximate the cost of the optimal path, i.e. $c(\pi) \approx c(\pi_*)$; (iv) number of iterations needed to reach the target node, as a measure of efficiency of the learnt heuristic function.

In our experiments, we compare A* run over learnt heuristics against a random baseline, sampling the heuristic values for nodes uniformly in [0, 1] and A* with {\it zero heuristics}, which is equivalent to running the Dijkstra's algorithm. Empirical analysis shows that the model exhibits very good out-of-distribution generalisation, retaining high accuracy on the constraints of \eqref{eq:objective} regardless of the dimension of the graph or the kind of connectivity, e.g. sparse, dense or very dense. This suggests that the model has learnt correctly how to output {\it consistent} heuristics, as the number of violated constraints is very low ($\leq$0.9\% on average, as shown in Table \ref{tab:results-t1}). Furthermore, the heuristic function can also lead A* to very fast convergence towards the target node. In fact, the number of iterations needed for A* to reach the target consistently outperforms Dijkstra and remains stable for all kind of graphs. Even for substantially larger graphs than the training ones, our "learnt" A* is greatly more efficient than running Dijkstra, as confirmed by the results on 256 nodes (around 15 iterations against 130 for very dense graphs). Also, the speedup metric measures the actual speedup in execution time (heuristic computation + search), guaranteeing that the overall time complexity of the learnt search does not exceed that of running classic Dijkstra.
In Fig. \ref{fig:results-dense} we show a comparison on dense graphs between A* and the random baseline for shortest path accuracy and relative distance, and a comparison among A*, Dijkstra and random baseline on the number of iterations. Here, we show the performance for all the available test sets, confirming good generalisation capabilities and fast convergence.

We can also witness interesting results on {\it sparse} graphs, where the model achieves the best performance with respect to relative distance. In this setting, the model might be helped by the fact that given a node, we can only route the path flow to a few nodes (given that the graph is sparser). Specifically, this may boost the relative distance metric as finding the optimal path "by accident" is more likely. However, the heuristics still leads to faster convergence towards the target node, indicating that the learnt heuristic function is still informative enough to guide the search towards that node. This also suggests that the model can generalise across different graph distributions, given that it has only been trained on {\it dense} graphs.

\section{Conclusions}
We have presented an unsupervised framework to learn heuristics for the A* search algorithm. Unlike existing approaches, we exploit neural algorithmic reasoning to bias the learning model to compute heuristics in an algorithmic fashion. The experimental analysis suggests that learning heuristics for A* through algorithmic reasoning and graph networks is doable and shows very good generalisation. We believe that our preliminary results should serve as a motivation for further work in this direction, extending the analysis to other search algorithms.



\bibliography{iclr2022_conference}
\bibliographystyle{iclr2022_conference}

\appendix
\label{appendix:appendix}

\section{Model selection} 
We employ a bi-level random search scheme for selecting the best hyperparameters for the model. 
The first random search draws uniformly the hyperparameter configurations for the hidden size in the interval [16, 512], learning rate in [1e-4, 1e-2], weight decay in [1e-5, 1e-1] and $\lambda$ in [1e-3, 1]. The second level refines the search and shrink the intervals to: hidden size in [80, 100], learning rate in [9e-4, 4e-3], weight decay in [9e4, 4e-3] and $\lambda$ in [5e-2, 1e-2].

\section{Model architecture}
In the following we specify the model architecture, e.g. layers and activations, to support reproducibility.
\begin{align*}
    \mathbf{z}_v &= f_v(\mathbf{x}_v ; \mathbf{\Theta}) = \mathbf{x}_v \mathbf{\Theta},  \\
    \mathbf{z}_{e} &= f_{e}(\mathbf{x}_{e} ; \mathbf{\Theta}) = \mathbf{x}_{e} \mathbf{\Theta},  \\
    \mathbf{h}^{(t)}_v &= P(\mathbf{z}_v, \mathbf{z}_e, \mathbf{h}_v^{(t-1)}) =\\
    &= 
        \psi_\theta\left([\mathbf{z}_v, \mathbf{h}_v^{(t-1)}], \mathsc{max} 
        \{ \phi_\theta([\mathbf{z}_v, \mathbf{h}_v^{(t-1)}], [\mathbf{z}_u, \mathbf{h}_u^{(t-1)}], \mathbf{z}_{e}) \mid \forall e \in E \,.\, e = (u, v) \}
        \right), \\
    \mathbf{o}^{(t)} &= g(\mathbf{z}_v, \mathbf{h}_v^{(t)}, \mathbf{h}_v^{(t-1)}  ; \mathbf{\Theta}) =
    [\mathbf{z}_v, \mathbf{h}_v^{(t)}, \mathbf{h}_v^{(t-1)}] \mathbf{\Theta}\,, \\
    \mathbf{y}^{(t)} &= g_h(\mathbf{z}_v, \mathbf{h}_v^{(t)}, \mathbf{h}_v^{(t-1)}  ; \mathbf{\Theta}) =
    [\mathbf{z}_v, \mathbf{h}_v^{(t)}, \mathbf{h}_v^{(t-1)}] \mathbf{\Theta}\,,
\end{align*}
where: $\mathbf{x}_v$,$\mathbf{x}_e$ refer to node/edge features; $[\cdot, \cdot]$ is concatenation; $\psi$ and $\phi$ are two neural networks with $\mathrm{ReLU}$ activations; $f_v$, $f_{e}$ are the encoders of node and edge features; and $g$, $g_h$ are the decoders of predecessors and heuristic node values.
\section{Loss function}
Here, we detail the loss function used to train our model:
\begin{align*}
    \mathcal{L}(\theta ; \lambda) = &\frac{1}{N} \sum_i^N \mathcal{L}_i(\theta ; \lambda), \\ 
\end{align*}
where $\mathcal{L}_i(\theta ; \lambda)$:
\begin{align*}
    \mathcal{L}_i(\theta ; \lambda) =
        &\frac{1}{T} \sum_t^T \mathcal{L}_{CE}(\mathbf{o}^{(t)}, \mathbf{p}^{(t)}) + \\
        &\frac{1}{T} \sum_t^T \big ( 
            (y^{(t)}_{s} - y^{(t)}_{g}) + 
            \sum_{(u,v) \in E} (y_v - y_u) \cdot \mathbf{1}_{[y_v - y_u > w_uv]} + \lambda ||\mathbf{y}^{(t)}||_2^2
        \big ).
\end{align*}
Here, $\mathcal{L}_{CE}$ is the cross-entropy function and $\mathbf{p}$ is the ground truth Dijkstra's output. $\mathbf{1}_{[y_v - y_u > w_uv]}$ is an indicator function checking satisfaction of a specific constraint, whereas $y_s$ and $y_g$ are the heuristic values for the source and target nodes respectively.

\end{document}